\documentclass[10pt, a4paper]{article}

\usepackage[final]{lrec2026}
\usepackage{amsmath}
\usepackage{tipa}

\title{AURORA Model of Formant-to-Tongue Inversion for Didactic and Clinical Applications}

\name{Patrycja Strycharczuk, Sam Kirkham} 

\address{University of Manchester, Lancaster University \\
         Manchester, UK, Lancashire, UK \\
         patrycja.strycharczuk@manchester.ac.uk, s.kirkham@lancaster.ac.uk}

\abstract{This paper outlines the conceptual and computational foundations of the AURORA (Acoustic Understanding and Real-time Observation of Resonant Articulations) model. AURORA predicts tongue displacement and shape in vowel sounds based on the first two formant values. It is intended as a didactic aid helping to explain the relationship between formants and the underlying articulation, as well as a foundation for biofeedback applications. The model is informed by ultrasound tongue imaging and acoustic data from 40 native speakers of English. In this paper we discuss the motivation for the model, the modelling objectives as well as the model architecture. We provide a qualitative evaluation of the model, focusing on selected tongue features. We then present two tools developed to make the model more accessible to a wider audience, a Shiny app and a prototype software for real-time tongue biofeedback. Potential users include students of phonetics, linguists in fields adjacent to phonetics, as well as speech and language therapy practitioners and clients.
 \\ \newline  }

\begin{document}

\maketitleabstract

\section{Introduction}
Vowel formants are a fundamental concept in speech science, because they reflect crucial auditory properties of vowel sounds. They are also relatively easy to measure and analyse, using popular speech analysis resources, such as Praat \citep{praat}. They are widely used for representing and comparing vowel systems in different languages and different varieties. In sociolinguistics, formant analysis helps track social and regional vowel changes \citep{labov2006}. In clinical linguistics, measuring formants can provide objective data to assess and diagnose selected speech disorders \citep{kent2003}. 

The above examples represent applications of formant analysis that are relevant to academic audiences who may have linguistic training, but not necessarily specialised phonetic background. Furthermore, there are situations where the notion of a formant has to be made accessible to lay audiences. \citet{kawitzky2020} propose a clinical intervention for voice feminisation based on formant biofeedback. The intervention is part of gender affirming voice training (GAVT), i.e. therapy helping transgender people produce speech that aligns with their gender identity. In contrast to many other applications of speech and language therapy, vowels are a prominent therapy goal in GAVT, as vowel formants have been shown to correlate with perception of speaker gender \citep{childers1991, leung2018, leyns2024}. More specifically, relatively higher formant values correlate with perceptions of femininity. To this end, \citet{kawitzky2020} 
use a real-time display of the LPC-spectrum to help a client achieve higher $F_2$ values in their speech. The client can observe the acoustic output of their own speech in real-time and is instructed by a therapist on how to achieve the change. \citet{mcallister2025} develop this approach further, presenting a web-version of the real-time LPC display with added functionality for helping the users adjust their resonances, including tutorials explaining the concept of resonance, and strategies for adjusting resonance.

A challenge that arises in this context is how to make the concept of a formant accessible to non-specialist audiences. Typically, this is done by interpreting formants in terms of the underlying articulation. \citet{kawitzky2020} advise their clients that moving the tongue forward in the mouth is a possible strategy for raising F2. This explanation is couched within a well-established articulatory metaphor for the interpretation of formants:  $F_1$ rises as the tongue lowers, and $F_2$ rises as the tongue advances. This articulatory interpretation of formant values is broadly correct, as it is fundamentally couched within the principles of the acoustic theory of speech production, which models acoustic vowel resonances as determined by the volume of the resonating cavities within the vocal tract whose size and shape is chiefly controlled by the tongue \citep{stevens1955, fant1971}. However, relating $F_1$ to tongue height and $F_2$ to tongue position without any further qualification involves considerable simplification. The generalisation leaves out several important factors: nonlinearity of resonance physics, acoustic coupling between $F_1$ and $F_2$, as well as interdependence between tongue height and shape. Due to biological properties of muscle movement, and the physical characteristics of resonance, tongue position and height are interdependent, as are $F_1$ and $F_2$. These properties are important to understanding the relationship between articulations and acoustics, but they also involve considerable conceptual complexity that cannot be fully addressed in clinical scenarios. Time is precious in a speech and language therapy clinic, and it is not feasible to present a client with a full account of biomechanical and acoustic properties of speech. 

The AURORA (Acoustic Understanding and Real-time Observation of Resonant Articulations) model is intended as a visual aid that bridges the connection between vowel formants and tongue movement in a way that balances simplicity and accuracy. It is an articulatory inversion model, predicting two-dimensional information about tongue displacement and tongue shape from input formant values. It is trained using synchronised ultrasound tongue images and audio recordings. It achieves a degree of articulatory accuracy thanks to modelling the interdependence of multiple articulatory dimensions and the interdependence of $F_1$ and $F_2$, informed by empirical data. It achieves simplicity by providing readily interpretable and low-dimensional visual output. We envision it as a didactic tool that can assist an explanation of articulation-acoustics relationships, as well as a foundation for the development of free and open-source tools for real-time biofeedback in vowel production.

\section{Training data}
The training data used to develop the model are synchronised ultrasound tongue imaging and audio data from 40 speakers. These data come from a pre-existing corpus. The recording and data processing procedures are described in detail in \cite{kirkham2023_co} and \cite{strycharczuk2025_dimensionality}.

 All the speakers were first-language speakers of English from the North of England. The age range was 18--48 (mean $=$ 24.6). 23 participants described their gender as female, 15 as male, one as non-binary, and one did not answer the gender question. 

The stimuli were real English words. They were all monosyllables including vowels within the b$\_\_$d segmental context and embedded within a fixed carrier phrase: \emph{bead,bird,bard,bored,booed,bid,bed,bad,bod,bud}. Note that British English is typically non-rhotic. The speakers produced between two and six repetitions (typically five) of the stimuli. The overall number of tokens was 3,856.

Ultrasound data were sampled at 59.5–101 frames per second (median = 81.3) and rotated to the speaker’s occlusal plane. Head stabilisation (UltraFit; \citealt{spreafico2018}) was used during recording to reduce probe movement. The ultrasound data were processed in Articulate Assistant Advanced Version 2.20 \citep{aaa}. The visible tongue contour in the ultrasound images was automatically tracked using DeepLabCut (DLC; \citealt{mathis2018}). We use a DLC model that identifies 11 consistent points along tongue from the epiglotic vallecula to the tongue tip \citep{wrench2022}. The coordinates of these points were extracted at the acoustic midpoint of the vowel. Corresponding $F_1$ and $F_2$ values at the same time point were also extracted. Formant measurements were obtained using FastTrack \citep{barreda2021}.

We performed further normalisation and dimensionality reduction on the tongue contour data in order to capture comparable information about the tongue across different speakers. Firstly we centered the data within speaker to reduce random variation related to differences in the ultrasound probe placement. From the centered data, we extracted information related to the position and height of the extreme points on the tongue contour: DLC knot 1, corresponding to the epiglottic vallecula, and DLC knot 11, corresponding to the tongue tip. These two dimensions allow us to capture the information about the position and rotation of the tongue contour on the original scale. Information about the tongue shape was captured using a morphometric approach described in \citet{dryden2016} and implemented using the {\tt shapes} package in R version 4.3.3. The approach involves a combination of Procrustes analysis and principal component analysis (PCA). Tongue contour coordinates are first translated, scaled and rotated using Procrustes analysis, so as to minimise variation in position, size or rotation of the tongue, preserving only variation in tongue shape. Normalised coordinates are then input into a PCA performed in the tangent space \citep{kent1992}. The first two PCs combined capture 67.54\% of variance in the data. We focus on these PCs only based on the previous findings on this type of PCA which show that the first two PCs capture most of the segmental variance between vowel categories, whereas the higher-order principal components mainly capture speaker-specific variance in tongue shape \citep{lo2025}. 

\section{The AURORA model}
The AURORA model is based on a multivariate linear model mapping the first two formants, $F_1$ and $F_2$, as well as an interaction between $F_1$ and $F_2$ (in Hz) onto six articulatory parameters, as described above: the height and position of the epiglottic vallecula (knot 1, in mm), the height and position of the tongue tip (knot 11, in mm) and two Principal Components related to tongue shape, as schematised in Equation~\ref{eq:model}.

\begin{equation}
\begin{bmatrix}
\text{knot 1 X} \\
\text{knot 1 Y} \\
\text{knot 11 X} \\
\text{knot 11 Y} \\
\text{PC1} \\
\text{PC2} \\

\end{bmatrix}
=
\beta_0
+ \beta_1 F_1
+ \beta_2 F_2
+ \beta_3 (F_1 \times F_2)
+ \varepsilon
\label{eq:model}
\end{equation}

The coefficients obtained from the training model are then used in the inversion procedure to predict the target articulatory parameters from any formant values input in Hz. 

\begin{figure}
    \centering
    a) PCA inversion to coordinate values
    \includegraphics[width=0.65\linewidth]{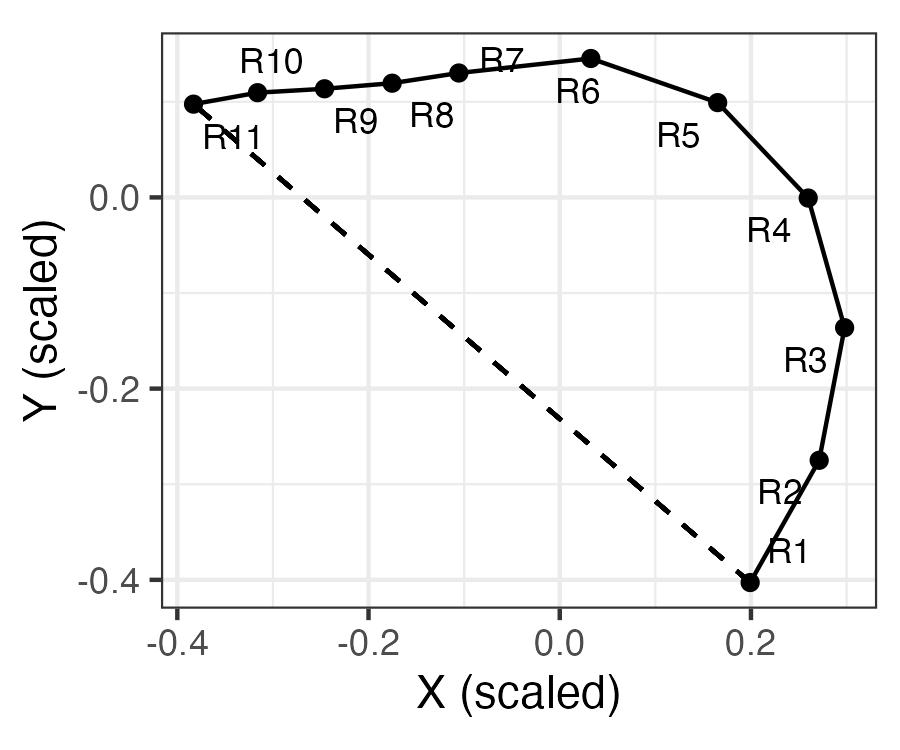}\\
    b) Reference vector
  \includegraphics[width=0.7\linewidth]{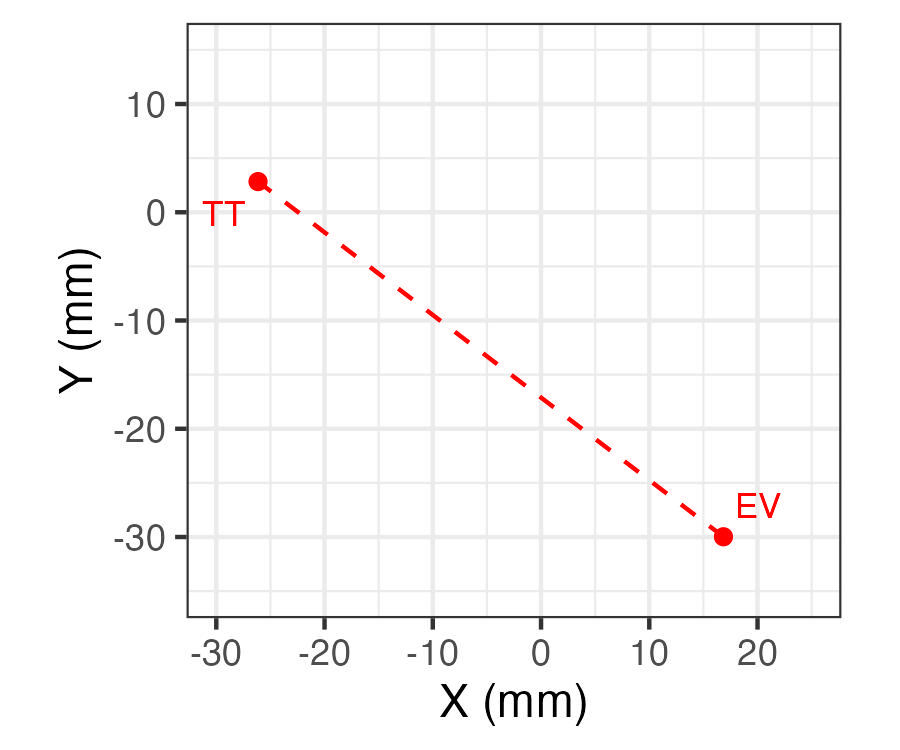}\\
       c) Rescaling, rotation and translation\\
         \includegraphics[width=0.7\linewidth]{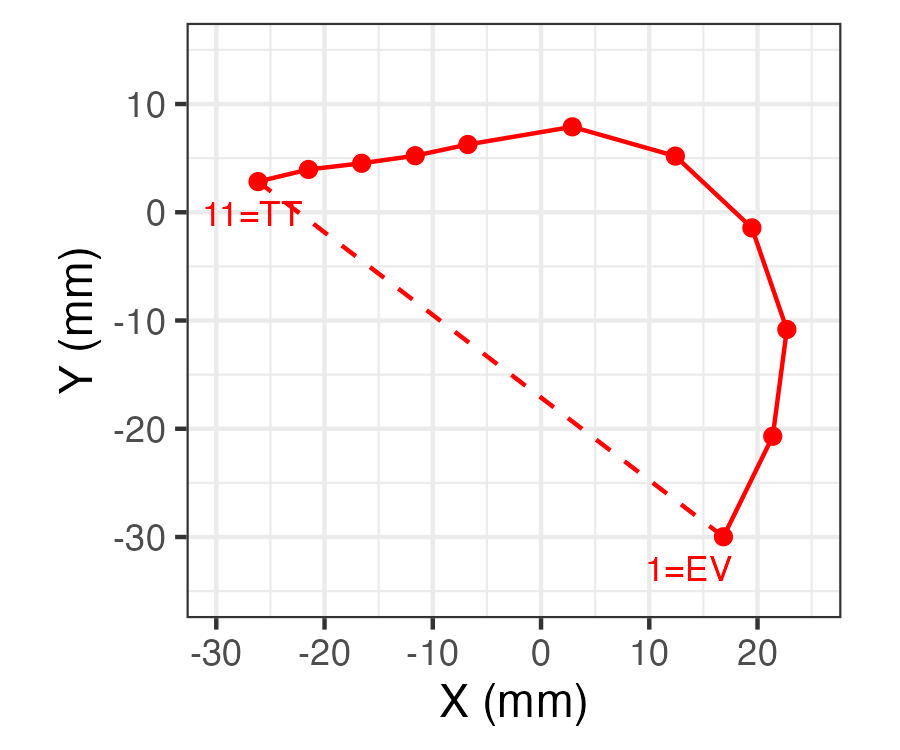}\\
      
                d) Smoothing\\
                       \includegraphics[width=0.7\linewidth]{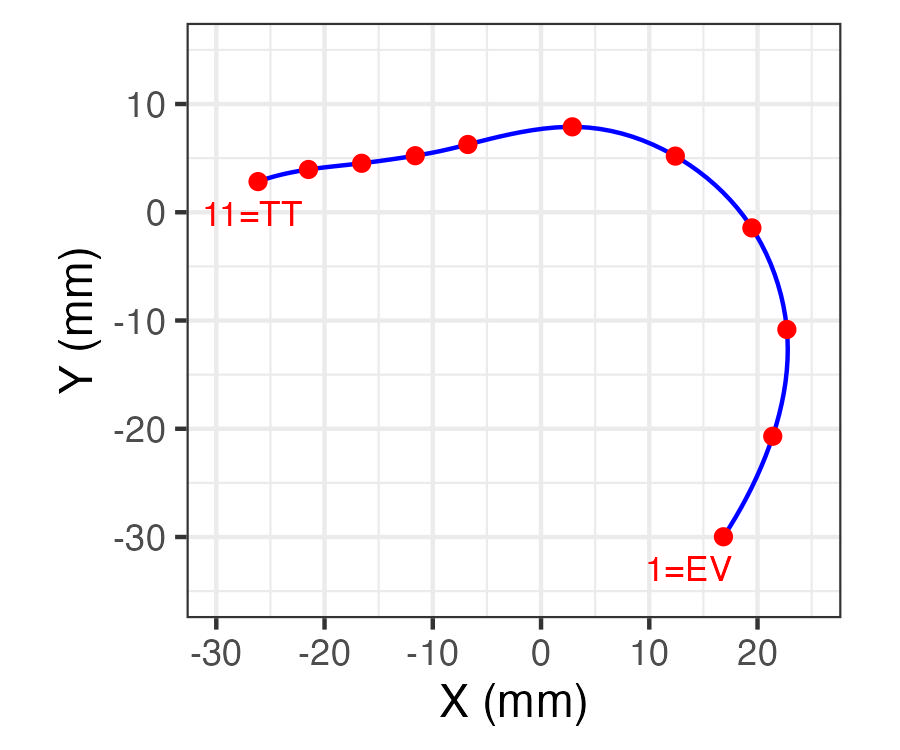}\\
    \caption{Stages of tongue contour reconstruction from the output of multivariate regression}
    \label{fig:rec}
\end{figure}

A separate procedure maps the regression predictors onto tongue contours, as illustrated in Figure \ref{fig:rec}. First, the PC values are used to reconstruct the tongue shape. The PCA is inverted by projecting the scores back through the eigenvector basis derived from the covariance structure of the tangent-space data. Thus obtained tongue shape (example in Figure \ref{fig:rec} a) is normalised, such that all predicted shapes are centred, scaled and rotated to the same plane. This normalisation is subsequently reversed during reconstruction stage, using Procrustes analysis. The vector between reconstructed knot R1 and reconstructed knot R11 (Figure \ref{fig:rec} a) is projected onto the reference vector between vallecula and tongue tip, as predicted by the multivariate model for the same set of input formant values. The reference vector is shown in Figure \ref{fig:rec} b. This determines the rotation, translation and scaling parameters, which are then applied to the reconstructed tongue coordinates. The output of this step is shown in (Figure \ref{fig:rec} c).

Finally, the data are smoothed and interpolated to span 100 points (from original 11). Smoothing is done, using piecewise cubic polynomials, as implemented in the {\tt smoothr} package \citep{smoothr}. The coordinates are interpolated independently, such that the coordinates of the DLC knots remain unchanged.  Figure \ref{fig:rec} d shows the output of the smoothing procedure, as well as the DLC knots. Note that it is not necessary to run this inversion process in real-time; e.g. Section 5 describes how the model can be used to generate a lookup table to enable fast inversion with low computational overhead.

\section{Evaluation}
The AURORA model focuses on generating useful tongue shape information for specific practical applications, and it does not aspire to be a maximally accurate model of articulatory inversion. It is deliberately limited by the small number of input parameters ($F_1$ and $F_2$), which helps illuminate the relationship between formants and the corresponding articulation. This is in line with the didactic goals of this project. With this in mind, we do not attempt to evaluate the model based on an objective measure of similarity between predicted articulation and a test set. Instead, we offer a qualitative evaluation guided by our original conceptual aims.

\begin{figure}
    \centering
    \includegraphics[width=1.2\linewidth]{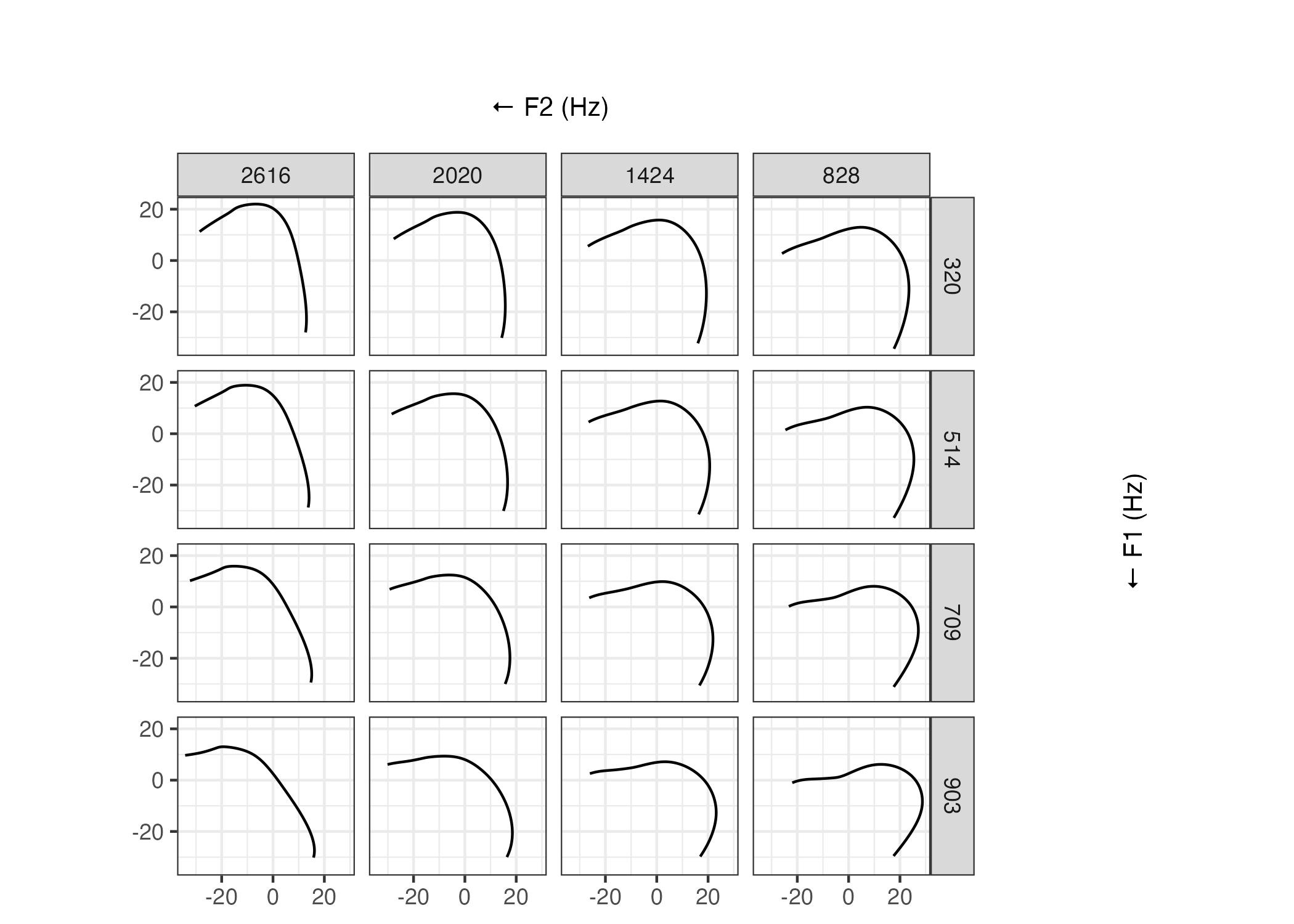}
    \caption{Tongue model predictions. Input $F_1$ and $F_2$ values are indicated on the strip labels. Tongue tip is on the left.}
    \label{fig:eval1}
\end{figure}

First, we evaluate the model through visual inspection of predicted tongue shapes for a range of representative $F_1$ and $F_2$ values. We generated a four-step continuum for $F_1$ and $F_2$ corresponding to the 5th and 95th quantile of these measures in our data. These quantiles correspond to 320 and 903 Hz for $F_1$, and 828 and 2616 Hz for $F_2$. We then used the model to predict tongue shapes for each possible combination of the $F_1$ and $F_2$ values from the four-step continuum. Figure \ref{fig:eval1} shows the tongue model predictions. As we can see from the figure, the model clearly differentiates the inherent tongue shape differences associated with differences in $F_1$ and $F_2$. Vowels with high $F_2$ are characterised by elongated and relatively straight tongue root, whereas vowels with low $F_2$ are characterised by convexity around the tongue dorsum and tongue root. $F_2$ also interacts with $F_1$ in affecting tongue shape, especially in the front part of the tongue. We find convexity there when $F_1$ is low and $F_2$ is high, and concavity when $F_1$ is high and $F_2$ is low. Associated with this systematic variation in the location of the highest point in the tongue, which is more anterior for vowels with high $F_2$, and more posterior for vowels with low $F_2$, with further interactions related to variation in $F_1$. Furthermore, the model captures differences in height variation, depending on $F_1$ and $F_2$ value. The same differences in $F_1$ correspond to greater variation in tongue height when $F_2$ is high, than when $F_2$ is low. 

\begin{figure}
    \centering
    \includegraphics[width=1.2\linewidth]{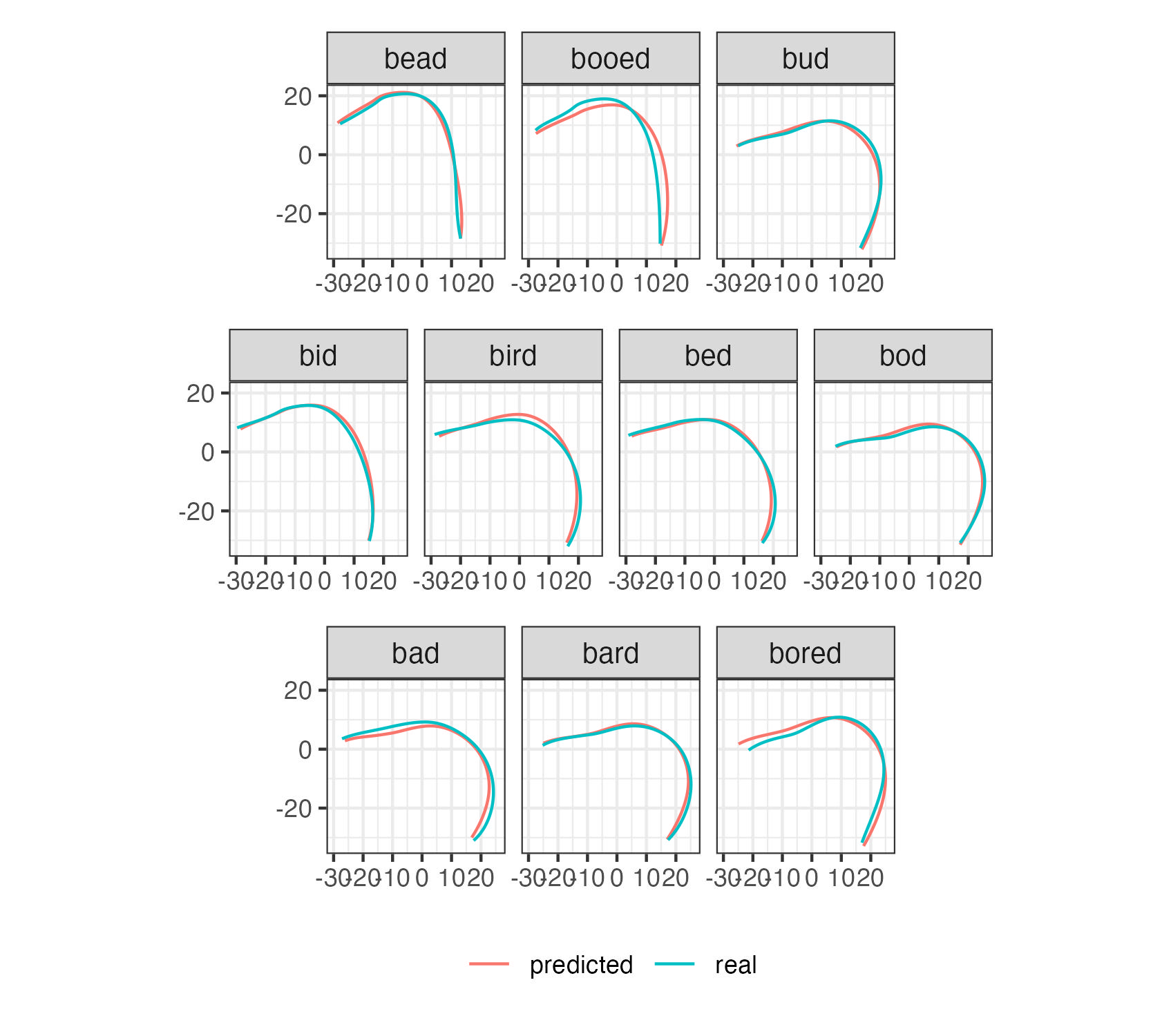}
    \caption{Comparison of tongue model predictions with mean tongue shapes for each item in the input data, based on the mean formant values. Tongue tip is on the left.}
    \label{fig:eval2}
\end{figure}

So far, we have shown that the model captures interactions between $F_1$ and $F_2$, as well as inter-dependencies between $F_1$, $F_2$, tongue position and tongue shape. These predictions are broadly as expected from existing models of vowel production \citep{maeda1990, honda1996, honda2010}. We further evaluate whether the model predictions are also realistic, by comparing the predictions to mean tongue data. We obtained the means by taking the by-item mean of each DLC knot across all speakers. Recall that the data were centered within speaker first. We then reconstructed the tongue contours from the DLC knots by interpolating the curve between each two neighbouring knots (similar as in Figure \ref{fig:rec} d). Model predictions were obtained based on mean $F_1$ and $F_2$ values (in Hz) for each item. The comparison between the real and predicted data is in Figure \ref{fig:eval2}.

In general, the model predictions are a very close match to the averaged tongue contours. The only case where the model and the mean tongue data are clearly mismatched is \emph{booed}. The tongue position predicted by the model is more retracted and lowered, compared to the mean data. This mismatch is easily explained by the fact that the model does not take lip rounding into account, and the vowel in \emph{booed} is rounded. Lip rounding has a lowering effect on formants, and without a specific correction, this leads the model to underestimate fronting and raising. We also find slight differences in tongue height between the model predictions and the mean tongue data for selected low vowels (\emph{bad, bored}). It is not entirely clear which aspects of the model architecture would cause this. The observed differences are relatively small and unlikely to lead to any major misinterpretation of the articulation-acoustics relationships.

In sum, the model provides a relatively rich and empirically accurate representation of tongue displacement and deformation corresponding to change in the first two formant values. It relies on common and well-understood analytical tools, such as linear regression, Procrustes analysis and principal component analysis. This makes the model entirely interpretable without much loss in accuracy. The simple tools are sufficient to capture the most pertinent biomechanical and acoustic facts relevant to explaining the general relationship between tongue position / shape and vowel formants. 

\section{Extensions and applications}
In order to make the model accessible to a wide audience, we have developed an interactive Shiny application\footnote{\url{https://aurora-model.shinyapps.io/tongue_shiny_app/}} that allows users to explore the data without programming. Users can adjust input parameters, using a slider that controls $F_1$ and $F_2$. The output is a visualisation of the predicted tongue contour, as well as the corresponding vowel spectrum and an indication of the vowel's location in the $F_1$$\sim$$F_2$ vowel space. The $F_1$$\sim$$F_2$ space is a common way of visualising formant results in linguistics. This interface is intended as a didactic aid for introductory phonetics teaching and for explaining the concept of formants to audiences with some linguistic training, but with no extensive background in speech science. The relevant audiences include academics in fields adjacent to phonetics, such as sociolinguistics, psycholinguistics, clinical linguistics and speech technology.

\begin{figure*}[!t]
    \centering
    \includegraphics[width=\textwidth]{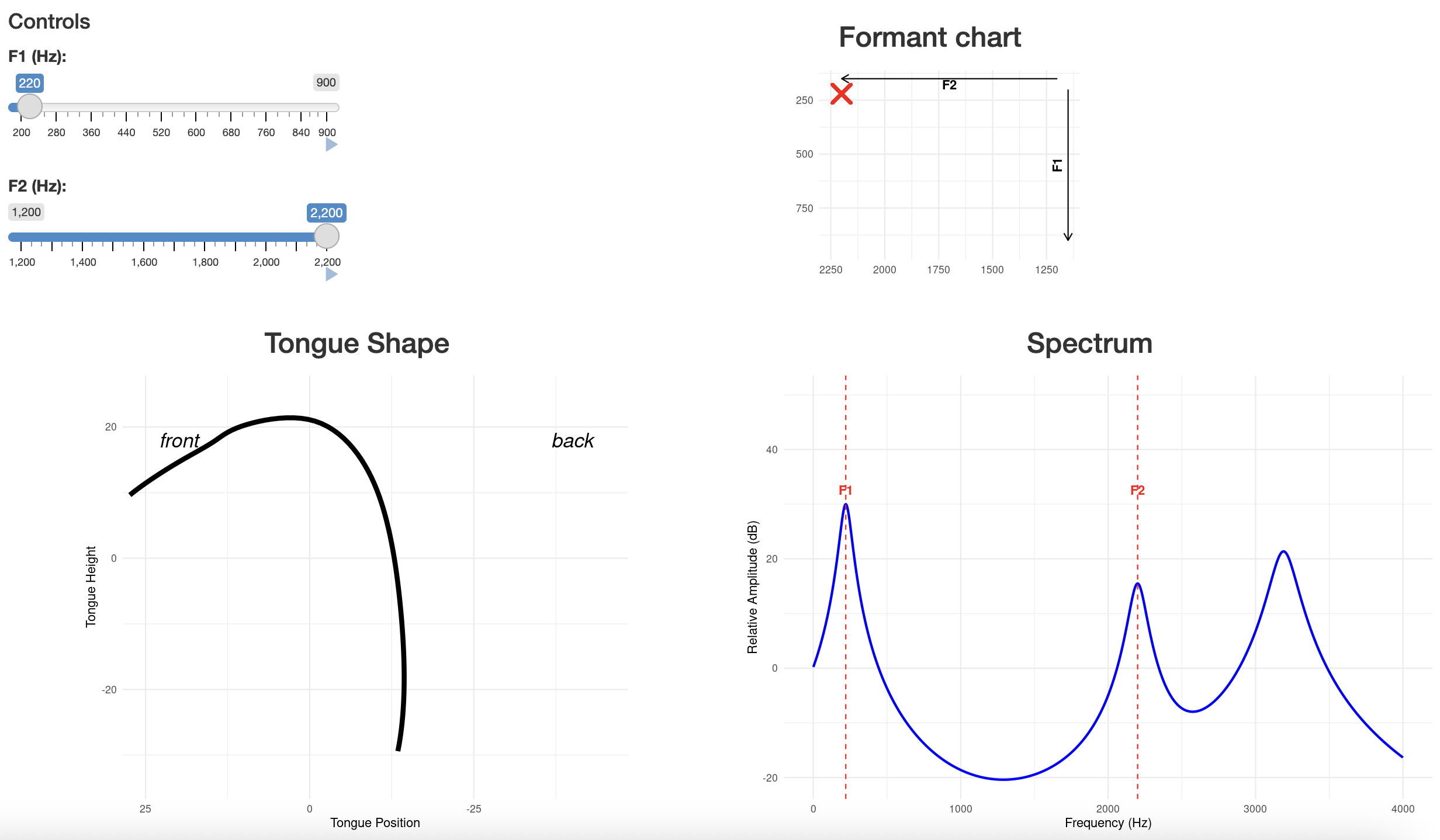}
    \caption{Screenshot of the Shiny app, showing the controls and the predicted output}
    \label{fig:shiny}
\end{figure*}

The other extension, which is currently still being developed, is a real-time biofeedback tool. This tool is inspired by the formant biofeedback model by \citet{mcallister2025} which visualises a stylised spectrum in real time, enhanced by real-time tracking of selected formants, which guides users towards target formant values, e.g. target $F_2$. Our biofeedback model is an extension of the same approach, as it displays the spectrum as well as the predicted tongue shape. The advantage of including a tongue contour is twofold. Firstly, the tongue contour is conceptually much simpler for the user to understand, compared to the notion of formant / resonance: it is immediate and does not require specific explanation. Secondly, the tongue display provides a bridge between the therapy goal (altering vowel resonances) and the instruction on how this might be achieved (changing tongue position / shape). From both of these perspectives, the inclusion of a tongue makes the biofeedback tool considerably more accessible to potential therapy clients. 

A preliminary version of the AURORA biofeedback software was implemented in Python 3.13 using PyQtGraph for the graphical user interface. The software analyses the incoming audio source in real-time and estimates the first four formant frequencies via Linear Predictive Coding (the temporal resolution of formant estimation can be set by the user). The estimated formants are passed in real-time via a pre-compiled look-up table that was constructed from the AURORA model described above. This generates a predicted tongue shape for each frame, which is displayed in real-time as a simple tongue contour.

\begin{figure*}[!t]
\centering
    \includegraphics[width=\textwidth]{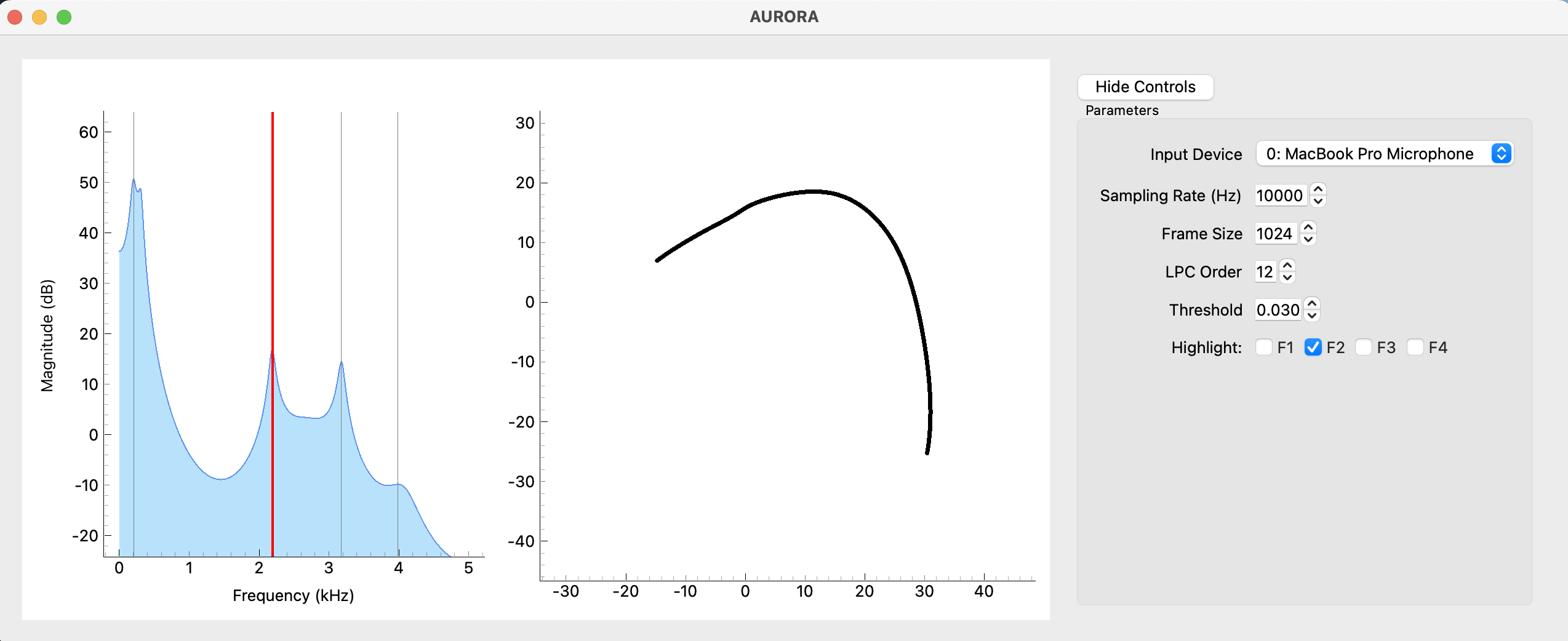}
    \includegraphics[width=\textwidth]{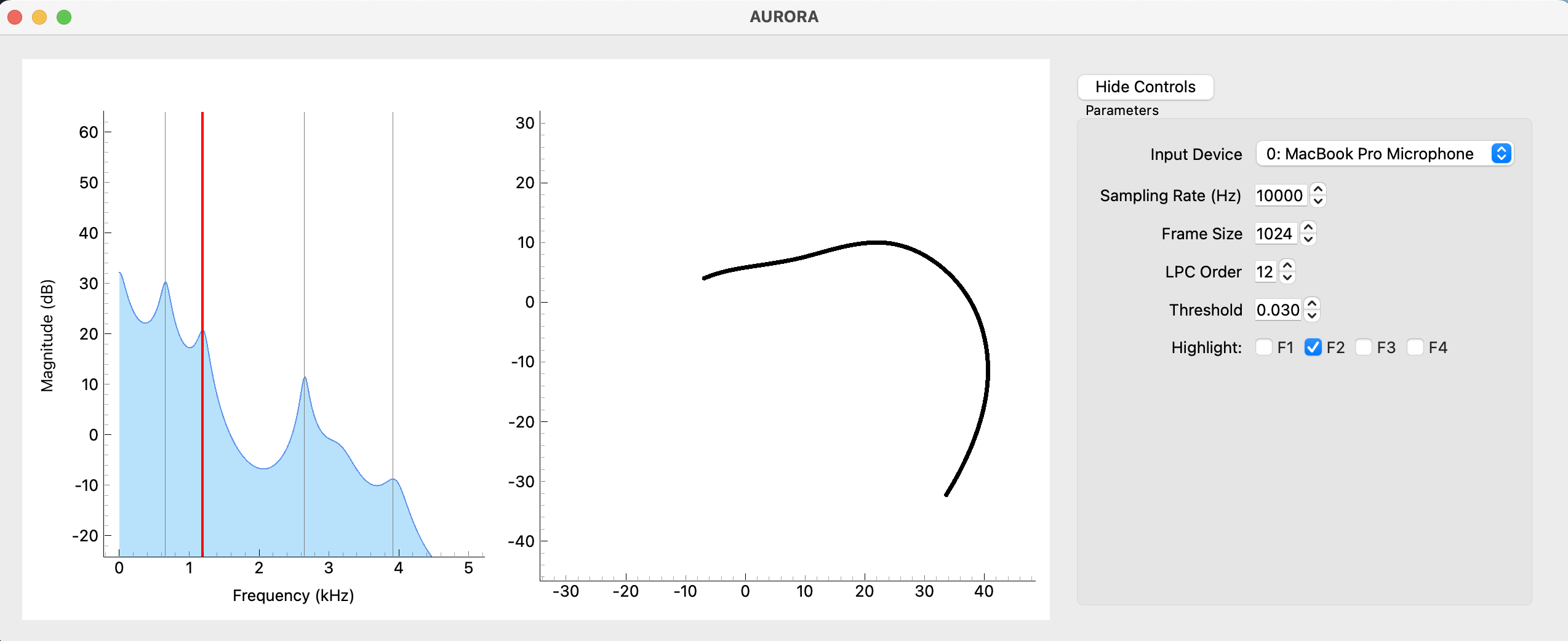}
    \caption{Screenshots of the acoustic/articulatory biofeedback app showing a production of /i/ (top) and /\textipa{A}/ (bottom). The left panel shows formant tracking based on the input spectrum and the middle panel shows a predicted tongue contour. The right panel shows controls, which can be hidden by the user.}
    \label{fig:biofeedback}
\end{figure*}

Example screenshots of the biofeedback tool are shown in Figure \ref{fig:biofeedback}. The top panel shows a single frame from a production of /i/ (e.g. \emph{beed}), with a predicted raising of the tongue dorsum, while the bottom panel shows \textipa{/A/} (e.g. \emph{bard}), with retraction of the tongue dorsum and tongue root. The user can highlight one or more formant peaks to help a client to focus on a particular formant -- in this case, we have highlighted $F_2$ -- which allows the user to focus on particular formants.

Alongside the spectrum and tongue display, the GUI also features a control panel, where the user can select the audio interface used for microphone input, as well as change settings such as sampling rate, frame size, LPC order, and an amplitude threshold above which formants are tracked. These settings allow the software to be adapted for different voices and in the future we hope to combine them into a more user-friendly calibration stage prior to using the tool.

\section{Limitations}

The most significant limitation of AURORA is that it only models the tongue. It can indirectly capture some information related to jaw displacement, based on horizontal displacement of the tongue, but it likely underestimates jaw movement, reflecting a more general limitation of ultrasound tongue imaging \citep{kirkham_jasael}. It does not correct for the involvement of other articulators that also affect formant values, such as the lips or the velum. It also does not contain any information about the passive articulators involved in vowel production. Resonances are determined by the distance between multiple articulators involved in the production of speech. As such, AURORA can adequately model selected key aspects of articulation involved in vowel production, but it does not fully visualise the way in which resonances are formed. 

Another potential limitation has to do with the fact that the model is trained on non-normalised formant values in Hertz. As such, it represents a mean between male and female formant values, which may not be representative of any typical speaker. Furthermore, given the greater proportion of female speakers in the training data, the model is possibly more representative of female speech. This does not impact the didactic applications of the model, but it is certainly an area for improvement for the development of future biofeedback tool which would rely on real-time formant tracking from individuals. The model can be easily adjusted to represent gender-specific formant values.

The model is trained on data from speakers of English from a specific locality. While the general principles of the articulation-acoustics relationship are universal, the model might possibly include some regional articulatory feature, reflecting dialect-specific articulatory setting. We have no specific evidence that this is the case, but we know that articulatory setting can be language- and dialect- specific \citep{laver1980, gold-2022}, so it is plausible that the model includes some specific bias. Inclusion of training data from other languages could illuminate this issue further, and greatly contribute to the educational value of this project.

\section{Conclusion}

We have outlined the developed of AURORA -- Acoustic Understanding and Real-time Observation of Resonant Articulations) -- a model for predicting tongue shape and movement from acoustic information. The model is simple and highly interpretable, because it uses a simple input feature set ($F_1$, $F_2$), well-understood statistical methods (linear regression, principal components analysis, Proctrustes analysis), and outputs interpretable tongue shape dimensions. We show that the model generates qualitatively accurate information on tongue shape that can directly inform biofeedback applications for speech and language therapy and phonetics teaching. Future work will focus on developing the visualisation and biofeedback tools, which can then be trialed for their efficacy in real-world clinical and phonetics training contexts.

\section{Acknowledgements}

The authors gratefully acknowledge support from the Arts and Humanities Research Council (AH/S011900/1 and AH/Y002822/1), the British Academy (MFSS24$\backslash$40076), and The Royal Society (APX\textbackslash R1\textbackslash 251102).

\section{Ethical considerations}

The study was exempt from ethical approval, as it constitutes secondary data analysis. The data collection for the original corpus received ethical approval and the participants provided informed consent, as reported in \citet{strycharczuk2025_dimensionality}.

We anticipate that the model and its applications should have primarily beneficial uses, such as improved speech and language therapy, accent training, and phonetics pedagogy. We emphasise that accent diversity is valuable and such tools should only be used by individuals who voluntarily seek accent training. These tools should not be used to pressure speakers to change their accent or to reinforce linguistic discrimination.

The model described in this paper was trained only using English data. While the model aims to capture universal physical principles of acoustic/articulatory relations and ought to generalise well to other languages, this has not yet been tested. As a result, caution should be exercised regarding the generality of these results.

\section{Bibliographical References}\label{sec:reference}
\vspace{-10mm}

\bibliographystyle{lrec2026-natbib}
\bibliography{biblio}

\end{document}